\documentclass[conference]{IEEEtran}
\IEEEoverridecommandlockouts
\usepackage{cite}
\usepackage{amsmath,amssymb,amsfonts}
\usepackage{algorithmic}
\usepackage{graphicx}
\usepackage{textcomp}
\usepackage{xcolor}
\def\BibTeX{{\rm B\kern-.05em{\sc i\kern-.025em b}\kern-.08em
    T\kern-.1667em\lower.7ex\hbox{E}\kern-.125emX}}
\begin{document}

\title{Predicting spatial distribution of Palmer Drought Severity Index*\\

\thanks{Identify applicable funding agency here. If none, delete this.}
}

\author{\IEEEauthorblockN{Vsevolod Grabar}
\IEEEauthorblockA{\textit{Skolkovo Institute of Science} \\
\textit{and Technology}\\
V.Grabar@skoltech.ru}
\and
\IEEEauthorblockN{Aleksandr Lukashevich}
\IEEEauthorblockA{\textit{Skolkovo Institute of Science} \\
\textit{and Technology}\\
a.lukashevich@skoltech.ru}
\and
\IEEEauthorblockN{Alexey Zaytcev}
\IEEEauthorblockA{\textit{Skolkovo Institute of Science} \\
\textit{and Technology}\\
A.Zaytsev@skoltech.ru}
}

\maketitle

\begin{abstract}
The probability of a drought for a particular region is crucial when making decisions related to agriculture. Forecasting this probability is critical for management and challenging at the same time. The prediction model should consider multiple factors with complex relationships across the region of interest and neighbouring regions. 

We approach this problem by presenting an end-to-end solution based on a spatio-temporal neural network. The model predicts the Palmer Drought Severity Index (PDSI) for subregions of interest. Predictions by climate models provide an additional source of knowledge of the model leading to more accurate drought predictions. 

Our model has better accuracy than baseline Gradient boosting solutions, as the $R^2$ score for it is $0.90$ compared to $0.85$ for Gradient boosting. Specific attention is on the range of applicability of the model. We examine various regions across the globe to validate them under different conditions. 

We complement the results with an analysis of how future climate changes for different scenarios affect the PDSI and how our model can help to make better decisions and more sustainable economics. 

\end{abstract}


\section{Introduction}
Why is it so important to monitor and forecast droughts? They are natural climate events that could happen in any type of terrain - starting from deserts and ending with tropical forests. At the same time, droughts are very costly, frequently happening, and affect the population and multiple economic sectors. Their impact is not limited by restricted geographical areas, such as riverbeds, coastlines or hurricane valleys. 

The development of drought is slow, while the consequences can be severe.
Thus, it is a good candidate for prediction on the base of past values on temperatures, precipitation, and overall conditions of soil and ground waters. To streamline the quantification of this event, practitioners use a multitude of indices~\cite{mishra2010review}. Among them, we consider, Palmer Drought Severity Index (PDSI)~\cite{alley1984palmer} and Selyaninov's Hydrothermal Coefficient (HTC)~\cite{chmist2022agricultural}, as they are relevant and easy to evaluate using available data and, thus, our methodology can be reused for other regions and conditions.

\paragraph{Palmer Drought Severity Index}
PDSI was developed in the 60s to monitor droughts that could affect the agricultural industry. It requires values of temperature, precipitation and soil water holding capacity. The soil water holding capacity could be replaced by default values in case of its non-availability. PDSI is widely used, and numerous examples of its applications (with code of PDSI estimation) can be found publicly. Finally, PDSI is not prone to seasonal problems. 
On the other hand, the estimation of this index requires the absence of missing data. PDSI is also lagging; sometimes, lags could reach up to 2 or 3 months, so we could not detect fast-evolving droughts. 

\paragraph{Selyaninov's Hydrothermal Coefficient}
HTC is an indicator widely used in Russia and tailored well to its climate conditions. It is rather straightforward to calculate, as it uses temperature and precipitation values:
\begin{equation}
    HTC = 10 \dfrac{\sum P_i }{ \sum T_i }
\end{equation}
where $\sum P_i$ and $\sum T_i$ are sums of precipitation and temperature over periods when the average temperature is over 10 degrees Celsius.
It is flexible enough to be used in both monthly and decadal applications. The main disadvantage of it is that HTC should be tuned to a specific region and it also doesn't use information about soil water.
HTC is also sensitive to dry conditions specific to the climate regime being monitored. 

\paragraph{Prediction}
While drought predictions and prediction of PDSI also happen in recent literature, the researchers focus on methods that ignore the spatio-temporal nature of the problem or aim at a selection of relevant features limiting the range of applicability of a prediction model to a single region~\cite{bacsakin2021drought}.

\paragraph{Contribution}
We managed to predict these indices using convolutional-temporal neural networks. They take into account both spatial and temporal dependencies in the data. Moreover, they can proceed large data volumes and don't require feature engineering for a specific problem.

\section{Data}
To evaluate our methods, we have used geospatial data that is available publicly - for example, Google Earth Engine~\cite{gorelick2017google}. 
To get the PDSI data, we employed TerraClimate Monthly dataset~\cite{abatzoglou2018terraclimate}. 
It has data for all of the Earth from 1958 up to 2021. 
For the calculation of HTC, we need daily data on precipitation and temperature, so we downloaded samples from the ERA5 collection~\cite{bell2021era5}. 

Extracted datasets are preprocessed. 
Our input looks like a 3d tensor, where one dimension is temporal (monthly), and two others are spatial (x and y coordinates of the grid). Datasets could be processed at various resolutions, so our grid dimensions vary from 9 by 16 and up to 40 by 200. 

\section{Methods}

We compare our approach with classic methods.
Among them, one of the most powerful is gradient boosting.

\subsection{Gradient boosting}

As a baseline, we consider gradient boosting of decision trees ensemble, which is implemented using the well-known library XGBoost~\cite{chen2016xgboost}. Treating every grid cell as a separate value, we narrow our task to a typical task of time series forecasting. XGBoost is both fast and efficient on a wide range of predictive modelling tasks and is a favourite among data science competition winners. 
It is an ensemble of decision trees, where new trees fix errors of those trees that are already part of the model. Trees are added until no further improvements can be made to the model.
Such an approach works well for geospatial data~\cite{proskura2019usage} and can deal with imbalanced problems~\cite{kozlovskaia2017deep} related to the drought prediction problem.

To improve our baseline and exploit the spatial information, we propose to use a modification called spatial gradient boosting. For this model, we add information about neighbouring cells.
For the sake of efficiency and simplicity of the model, we consider a 3x3 neighbourhood. So, we include eight additional time series. 
For "edge" cells, some of them are complete zeros. 

\subsection{ConvLSTM}
Our model is inspired by paper~\cite{kail2021recurrent} and is a version of Covolutional LSTM architecture from~\cite{shi2015convolutional}. 
We adopt recurrent neural networks to reflect temporal dependencies. 
In particular, we use LSTM, a type of RNN which uses an additional state cell to enable long-term memory. There we employ two-dimensional feature maps as hidden states compared to one-dimensional hidden states in common LSTM, as we require grid-to-grid transformations. Next, we process a grid of PDSI values for various grid dimensions, from 9x16 to 40x200. 
To benefit from spatial dependencies between drought severity values, we use convolution neural networks, which efficiently work with images and other two-dimensional signals like geospatial data. Our approach combines RNN and CNN architectures as we pass information through RNN in the form of a feature map obtained using CNN.

\subsection{Our architecture}

\begin{figure}[h]
    \centering
    \includegraphics[scale=0.35]{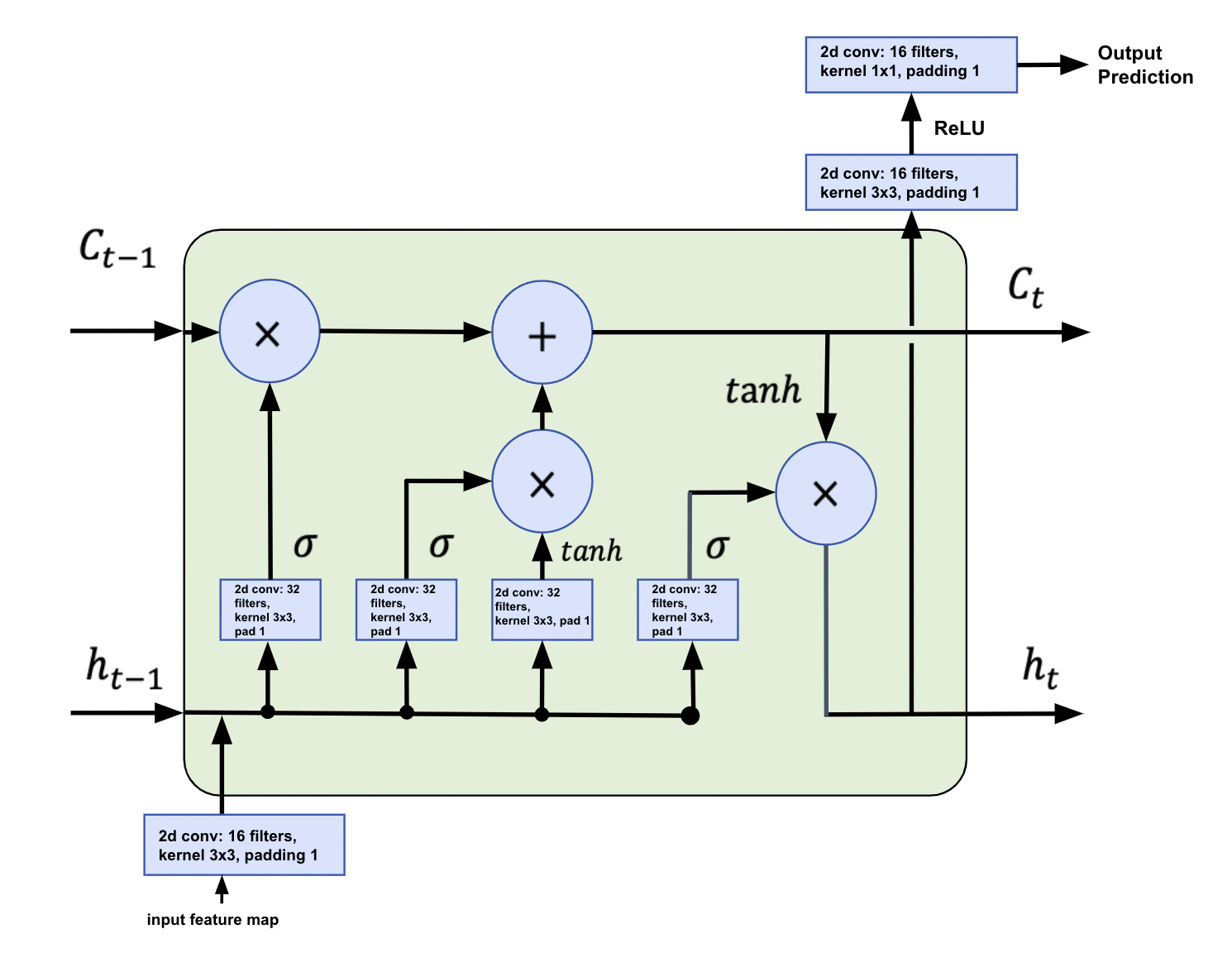}
    \caption{The proposed version of ConvLSTM architecture}
    \label{fig:convlstm}
\end{figure}

Our main architecture depicted in Figure~\ref{fig:convlstm} follows the pipeline:
\begin{enumerate}
    \item  We represent data as a sequence of grids: for each cell, we specify a value of float drought index on a particular month;  the input grid at each time moment has various dimensions (e.g. 9x16 or 40x200) 
    \item We pass the input grid through a convolutional network to create an embedding of grid dimensionality size with 16 channels. As an output of LSTM at each time moment, we have a hidden representation (short term memory) of size $32\times grid_h \times grid_w$, cell (long term memory) representation of a similar size, and the output of size $32\times grid_h \times grid_w$. 
    \item We transform the output to the size $1\times grid_h \times grid_w$ using convolution $1 \times 1$ to receive values of our indicator for each cell as a final prediction. As an additional hyperparameter, we vary the forecasting horizon - i.e. we can forecast PDSI for the next month or $k$-th month.
\end{enumerate}

\section{Results}

\subsection{Evaluation procedure}

For evaluation of the model, we use $R^2$ score.
Higher values of $R^2$ scores correspond to better models, with the perfect value being $1$ and the value for a random prediction being around $1$.

\subsection{Main results}

We have observed that our model performed better than the baseline gradient boosting tree algorithm, reaching $R^2$ score of $0.9$ as compared with a common Gradient boosting score of $0.85$ and a spatial Gradient boosting score of $0.85$.

\subsection{Prediction and errors for a particular region}

\begin{figure}[!h]
    \centering
    \includegraphics[scale=0.50]{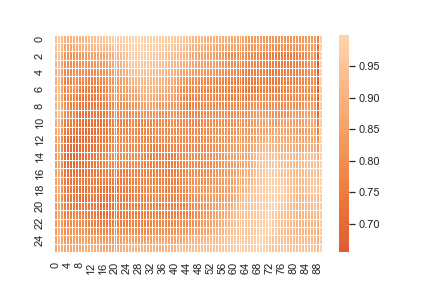}
    \caption{Spatial distribution of $R^2$, Iowa state, ConvLSTM}
    \label{fig:r2_spatial}
\end{figure}

We have taken the approximate region of Iowa state to track the performance of our algorithm. 
The distribution of $R^2$ scores is at Figure~\ref{fig:r2_spatial}.

We observe that the distribution of $R^2$ is non-uniform across the cells: the standard deviation is quite large, and individual values vary from negative ones to perfect, equaling almost $1.0$. 

\begin{figure}[!h]
    \centering
    \includegraphics[scale=0.50]{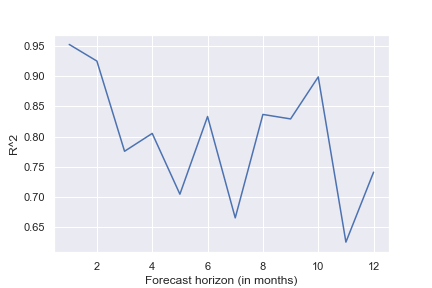}
    \caption{Dynamics of $R^2$ with respect to forecast horizon}
    \label{fig:forecast_horizon}
\end{figure}

As a sanity check, we also study how average $R^2$ changes with the increase of the forecasting horizon.
The dynamic of $R^2$ scores is given in Figure~\ref{fig:forecast_horizon}.

As expected, longer prediction horizons lead to lower model quality.
While the 1-month look ahead is close to $1$ being approximately $0.95$, it gradually falls to $0.75$ at the 6-month horizon.

\section{Conclusion}

Droughts are severe natural disasters affecting the economy through various channels, including agriculture and population well-being. Due to climate change, droughts happen more frequently around the world, and their amplitude has also increased, which could be witnessed from the summer of 2022 in the Northern Hemisphere. 

Thus, predicting future droughts accurately and preparing for their effects would help to mitigate some of the negative effects of climate change. We have observed from our research study that combining convolutional neural nets with recurrent ones significantly increased the forecasting abilities of various drought indicators (as compared with previous algorithms), including PDSI. 

\bibliographystyle{plain}
\bibliography{main.bib}

\end{document}